\documentclass{article}

\usepackage{microtype}
\usepackage{graphicx}
\usepackage{subfigure}
\usepackage{booktabs} 
\usepackage{tikz} 
\usetikzlibrary{arrows,calc,fit, patterns}
\definecolor{blue}{rgb}{0.5,0.5,0.5}
\definecolor{tegreen}{RGB}{21,180,113}
\usetikzlibrary{arrows.meta}

\usepackage{hyperref}

\usepackage{todonotes}

\usepackage[accepted]{icml2019}

\icmltitlerunning{Continual Reinforcement Learning deployed in Real-life using Policy Distillation and Sim2Real Transfer}

\begin{document}

\twocolumn[
\icmltitle{Continual Reinforcement Learning deployed in Real-life using Policy Distillation and Sim2Real Transfer}



\icmlsetsymbol{equal}{*}

\begin{icmlauthorlist}
\icmlauthor{Ren\'e Traor\'e}{equal,ensta}
\icmlauthor{Hugo Caselles-Dupr\'e}{equal,ensta,softbank}
\icmlauthor{Timoth\'ee Lesort}{equal,ensta,thales}
\icmlauthor{Te Sun}{ensta}
\icmlauthor{Natalia D\'iaz-Rodr\'iguez}{ensta}
\icmlauthor{David Filliat}{ensta}
\end{icmlauthorlist}

\icmlaffiliation{ensta}{Flowers Laboratory (ENSTA ParisTech, INRIA)}
\icmlaffiliation{softbank}{AI Lab, Softbank Robotics Europe}
\icmlaffiliation{thales}{Theresis Lab, Thales}

\icmlcorrespondingauthor{Hugo Caselles-Dupr\'e}{caselles@ensta.fr}
\icmlcorrespondingauthor{Timoth\'ee Lesort}{timothee.lesort@ensta.fr}
\icmlcorrespondingauthor{Ren\'e Traor\'e}{rene.traore@ensta.fr}

\icmlkeywords{Machine Learning, ICML}

\vskip 0.3in
]



\printAffiliationsAndNotice{\icmlEqualContribution} 

\begin{abstract}
We focus on the problem of teaching a robot to solve tasks presented sequentially, i.e., in a continual learning scenario. The robot should be able to solve all tasks it has encountered, without forgetting past tasks. We provide preliminary work on applying Reinforcement Learning to such setting, on 2D navigation tasks for a 3 wheel omni-directional robot. Our approach takes advantage of state representation learning and policy distillation. Policies are trained using learned features as input, rather than raw observations, allowing better sample efficiency. Policy distillation is used to combine multiple policies into a single one that solves all encountered tasks.
\end{abstract}

\section{Introduction}

In realistic real-life reinforcement learning scenarios, involving for instance service robots, tasks evolve over time either because the context of one task changes or because new tasks appear \cite{Doncieux18}. Our end goal is therefore to have an embodied agent in real-life that learns incrementally as time passes. One example would be a robot tasked with wrapping gifts. Most gifts are rectangular packages (cuboids), so the robot would first learn to wrap cuboids. Then if a soccer ball appears, the robot would have to learn how to wrap a sphere while still being able to wrap cuboids. Indeed, the robot should add additional knowledge to his repertoire. Moreover, even if it would be easier to learn how to wrap spheres and cuboids before test time, there are potentially many other shapes that have to be considered, and thus, learning continually seems more natural and convenient than trying to learn all skills at once. 

Continual Learning (CL) and State Representation Learning (SRL) are essential to build agents that face such challenge. 
SRL allows to build strong representations of the world since agents should be able to understand their surroundings, and extract general concepts from sensory inputs of complex scenes. An agent better sees a chair as an object, not as a bunch of pixels together in an image.
CL allows to learn such representation without forgetting in settings where the distribution of data change through time and is needed for agents that learn in the real-world and are required to adapt to changes. Combining CL and SRL would then allow to create strong representation robust to catastrophic forgetting.

\begin{figure}
    \centering
    \includegraphics[scale=0.2]{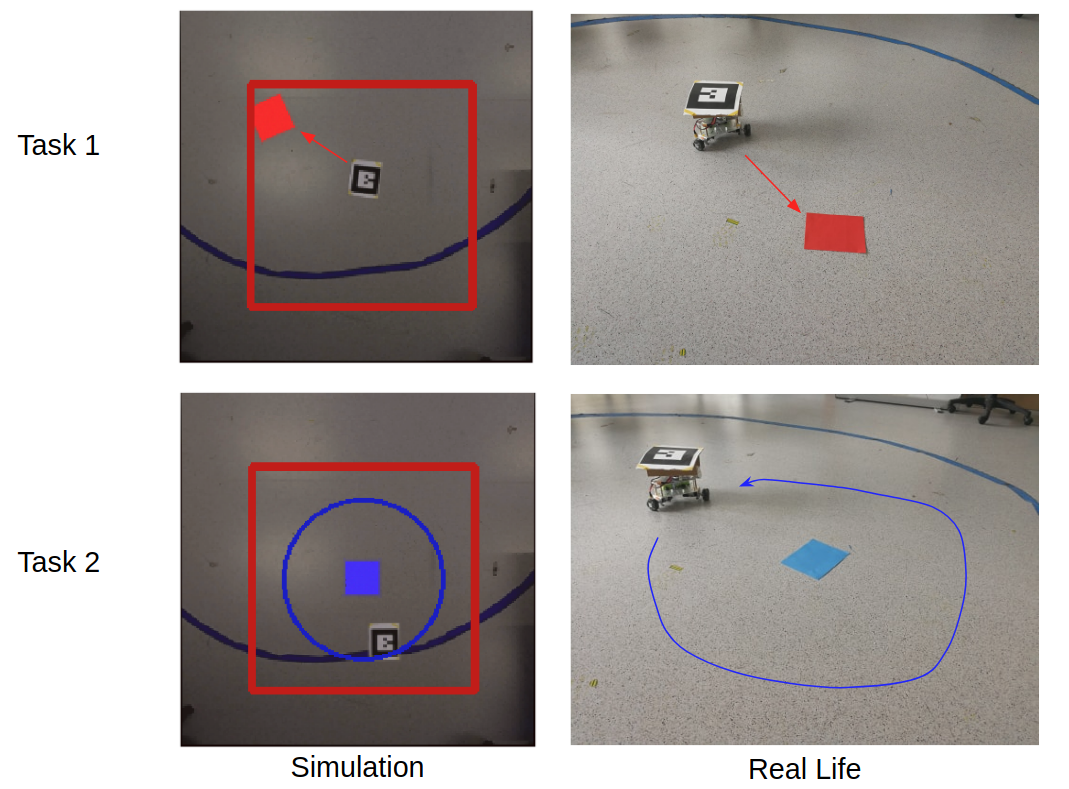}
    \caption{Having access to task 1 only first, and then task 2 only, we learn a single policy that solves two real-life navigation tasks using policy distillation and sim2real transfer.}
    \label{fig:real_life_tasks}
\end{figure}

Reinforcement learning (RL) is a popular approach to learn robot controllers that also has to face the CL challenges, and can take advantage of SRL to learn faster and to produce more robust policies. Therefore we perform our experiments (Fig. \ref{fig:real_life_tasks}) in a setup where tasks are encountered sequentially and not all at once. Note that it differs from a setting where we can pick and shuffle experiences, often encountered in the multi-task RL literature (cf Section \ref{subsec:multi}).

In our approach we aim to take advantage of simulations to create this scenario. We demonstrate 
that deploying a policy in real-life which has continually learned two tasks in simulation is successful with our approach.

Our contribution consists on applying two major paradigms for robotics in real life: a state representation learning approach for compact and efficient representation that facilitates learning a policy, and a policy that learns continually in a sequential manner. The approach is deployed in a real robot thanks to policy distillation and sim2real transfer. Furthermore, in opposition to most methods in reinforcement learning, the solution we propose does not need a task indicator at test time. Indeed, the information about the task to be solved can be found from a different color tag in the image.

The rest of the article is structured as follows. Section \ref{ref:relatedwork} introduces state representation learning, multi-task RL and continual learning paradigms in an RL setting, Sec.\ref{ref:experiments} details the robotics settings and tasks performed; Sec.\ref{ref:methods} details the methods utilized and Sec.\ref{ref:discussion} concludes with future insights from our experiments.

\section{Related work}
\label{ref:relatedwork}

\subsection{State representation learning (SRL)}

Scaling end-to-end reinforcement learning to control real robots from vision presents a series of challenges, in particular in terms of sample efficiency. Against end-to-end learning, SRL \cite{Lesort18} can help learn a compact, efficient and relevant representation of states. Previous works such as \cite{Finn15, Watter15, Hoof16,Lesort17, Sermanet17, Thomas17, raffin2019decoupling} have shown that SRL can speed up policy learning, reducing the number of samples needed while additionally being easier to interpret. 


\subsection{Multi-task RL}
\label{subsec:multi}

Multi-task RL aims at constructing one single policy module that can solve a number of different tasks. The CURIOUS algorithm \cite{colas2018curious} selects through exploration the tasks to be learned that improve an absolute learning progress metric the most. 
Policy distillation \cite{rusu2015policy} can also be used to merge different policies into one module/network. The Distral algorithm \cite{teh2017distral} is one successful example of such approach: a shared policy distills common behaviours from task-specific policies. Then, the distilled policy is used to guide task-specific policies via regularization using a Kullback-Leibler (KL) divergence.
Other approaches like SAC-X \cite{riedmiller2018learning} or HER \cite{andrychowicz2017hindsight} take advantage of Multi-task RL by learning auxiliary tasks in order to help the learning of an objective task. 

\subsection{Continual Learning}

Continual learning (CL) is the ability of a model to learn new skills without forgetting previous knowledge. 
In our context, it means learning several tasks sequentially and being able to solve any task at the end of the sequence. This differs from the easier multi-task scenario, where tasks can be experienced all at once. 

Most CL approaches can be classified into four main methods that differ in the way they handle the memory from past tasks. 

The first method, referred to as \textit{rehearsal}, keeps samples from previous tasks 
\cite{rebuffi2017icarl, nguyen2017variational}.
The second approach consists in applying \textit{regularization}, either by constraining weight updates in order to maintain knowledge from previous tasks 
\cite{kirkpatrick2017overcoming, Zenke17, Maltoni18}, or by keeping an old model in memory and distilling knowledge \cite{hinton2015distilling} into it later to remember \cite{li2018learning, schwarz2018progress, rusu2015policy}.
The third category of strategies, \textit{dynamic network architectures}, maintains past knowledge thanks to architectural modifications while learning \cite{rusu2016progressive, fernando2017pathnet, li2018learning, fernando2017pathnet}.
The fourth and more recent method is \textit{generative replay} \cite{shin2017continual, lesort2018generative, wu2018memory, lesort2018marginal}, where a generative model is used as a memory to produce samples from previous tasks. This approach has also been referred to as pseudo-rehearsal.
Note that all four type of methods can used for classification as well as for generation.


\subsection{RL in real-life}

Applying RL to real-life scenarios is a major challenge that has been studied widely. Most attempts fall into two categories: games and robotics.

For games, AlphaGo Zero \cite{silver2018general} has mastered the game of Go from scratch without any human supervision by combining RL, self-play and Monte Carlo Tree Search \cite{chaslot2008monte}. AlphaStar \cite{alphastar} and OpenAI Five \cite{OpenAI_dota} were both able to get competitive results against professional human players on the game Starcraft and DOTA2, respectively. Both solutions are based on RL, and current research is still investigating how to master the game with the same constraints as humans (e.g. same FPS). 

In robotics, there is a plethora of successful attempts at deploying RL on real robots. One common approach is training policies in simulation and then deploying them in real-life hoping that they will successfully transfer, considering the gap in complexity between simulation and the real world. Such approaches are termed \textit{Sim2Real} \cite{Golemo19}, and have been successfully applied \cite{christiano2016transfer, matas2018sim} in many scenarios. In order to cope with the unpredictable nature of the real world, one can use Domain Randomization \cite{tobin2017domain}, which we use in our approach. This technique trains policies in numerous simulations that are randomly different from each other (different background, colors, etc.). Using this technique, the transfer to real life is easier.

Others have tried to train a policy directly on real robots, facing the hurdle of the lack of sample efficiency that RL suffers from. SAC-X \cite{riedmiller2018learning} is one example where a successful policy is learned directly on the real robot. 

One can also find applications of reinforcement learning in other domains: TacTex'13 \cite{AAAI14-urieli}, relying on online RL, is an autonomous broker agent that maximizes profit through energy trading;  and \cite{DBLP:journals/corr/LiMRGGJ16} propose using policy gradients for dialogue generation using a set of reward functions designed to increase the diversity and length of generated responses. In education, a faster teaching policy by {{POMDP}} (partially-observable
Markov decision process) planning \cite{Rafferty2016FasterTV} leverages a probabilistic learner model in order to achieve a long-term teaching objective.

In the literature, most approaches focus on the single-task or simultaneous multi-task scenario. In this paper, we attempt to train a policy on several tasks sequentially and deploy it in real life. Hence, we attempt to apply RL in real life in a continual learning setting.


\section{Methods}
\label{ref:methods}


In this section we present the method proposed to combine state representation learning (SRL) and continual learning (CL) in a real life reinforcement learning setting. First we present how a single task is learned and how the SRL part works, secondly we explain how to learn continually and thirdly, we explain how we evaluate learning in the different phases of the learning sequence.

\subsection{Learning on one task}
\label{subsec:oneTask}

\usetikzlibrary{arrows}
\usetikzlibrary{decorations.markings}
\newcommand{\mygrid}{\tikz{\draw[step=0.5cm] (0,0)  grid (0.5,1.5);}}

\begin{figure}[ht!]
\centering

\resizebox{0.45\textwidth}{!}{
\begin{tikzpicture}[
roundnode/.style={circle, draw=black!60, fill=green!0, very thick, minimum size=10mm},
roundnode2/.style={circle, draw=black!60, fill=black!20, very thick, minimum size=10mm},
squarednode_st/.style={rectangle, draw=black!60, fill=green!20, very thick, minimum size=10mm},
squarednode/.style={rectangle, draw=black!60, fill=black!20, very thick, minimum size=10mm},
squarednode_st2/.style={rectangle, draw=black!60, very thick, minimum width=10mm, minimum height = 2cm},
invisible/.style={rectangle , draw=black!0, fill=green!0, very thick, minimum size=10mm},
squarednode_img/.style={rectangle, draw=black!60, fill=black!20, very thick, minimum size=10mm},
container_AE/.style={draw, rectangle, draw=green!60, dashed, inner sep=1em},
container_Inv/.style={draw, rectangle, draw=blue, dashed, inner sep=1em},
]


\node[invisible]        (base)        {};
\node[invisible]        (base2)        [above=of base] {};

\node[invisible]        (obs)        {};
\node[invisible]        (obs2)        [above=of obs] {};

\node[roundnode]        (hidden_obs2)        [on grid,above=of base2] {$I_{t+1}$};
\node[roundnode]        (hidden_obs)        [on grid,below=of base] {$I_t$};
\node[invisible]        (hidden_obs3)        [on grid,above=of base] {};



\node[invisible]        (hidden_state)        [right=of hidden_obs3,draw] {};
\node[invisible]        (hidden_state)        [on grid,right=of hidden_state,draw] {};

\node[squarednode_st2] (anode) [right=of hidden_obs,draw, pattern=horizontal lines light gray]{};
\node[] (label) [on grid, above=0.8cm of anode]{$s_{t}$};

\node[squarednode_st2] (anode2) [right=of hidden_obs2,draw, pattern=horizontal lines light gray]{};
\node[] (label2) [on grid, above=0.8cm of anode2]{$s_{t+1}$};


\node[invisible]        (center3)        [right=of hidden_state] {};

\node[squarednode]        (recon_act)        [on grid, above=of center3, draw] {$\hat{a}_t$};
\node[squarednode]        (recon_img)        [on grid,below=of center3,draw] {$\hat{I}_t$};

\node[roundnode]        (img)        [right=of recon_img,draw] {$I_t$};
\node[roundnode]        (act)        [right=of recon_act,draw] {$a_t$};

\node[invisible]        (L_img)        [right=of img,draw] {$\mathcal{L}_{Reconstruction}$};
\node[invisible]        (L_act)        [right=of act,draw] {$\mathcal{L}_{Inverse}$};


\draw[-{Latex[length=3mm,width=2mm]}] (hidden_obs.east) -- (anode.west);
\draw[-{Latex[length=3mm,width=2mm]}] (hidden_obs2.east) -- (anode2.west);


\draw[-{Latex[length=3mm,width=2mm]}] (anode2.east) -- (recon_act.west);
\draw[-{Latex[length=3mm,width=2mm]}] (anode.east) -- (recon_act.west);

\draw[-{Latex[length=3mm,width=2mm]}] (anode.east) -- (recon_img.west);

\node[container_AE, fit=(recon_act) (act)] (fwd) {};
\node[container_Inv, fit=(recon_img) (img)] (ae) {};

\end{tikzpicture}
}

\caption{\textit{SRL Combination} model: combines the prediction of an image $I$'s reconstruction loss and an inverse dynamic model loss in a state representation $s$. Arrows represent inference, dashed frames represent losses computations, rectangles are state representations, circles are real observed data, and squares are model predictions, $t$ represents the timestep}

\label{fig:split-model}
\end{figure}
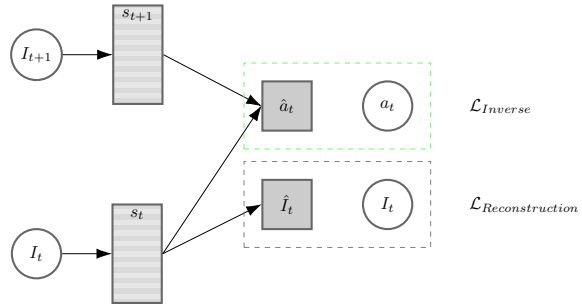


Each task $i$ is learned following to procedure we describe here.
First, as we use an SRL approach, we need to learn a state representation encoder.
We sample data from the environment $Env_i$ (where $i$ refers to the task) with an agent guided by a random policy. We call this dataset $D_{Random,i}$.
$D_{Random,i}$ is then used to train an SRL model composed of an inverse model and an auto-encoder. This architecture is inspired from \cite{raffin2019decoupling}, and illustrated in Fig. \ref{fig:split-model}.


\begin{figure*}
\centering
\includegraphics[width=.9\textwidth]{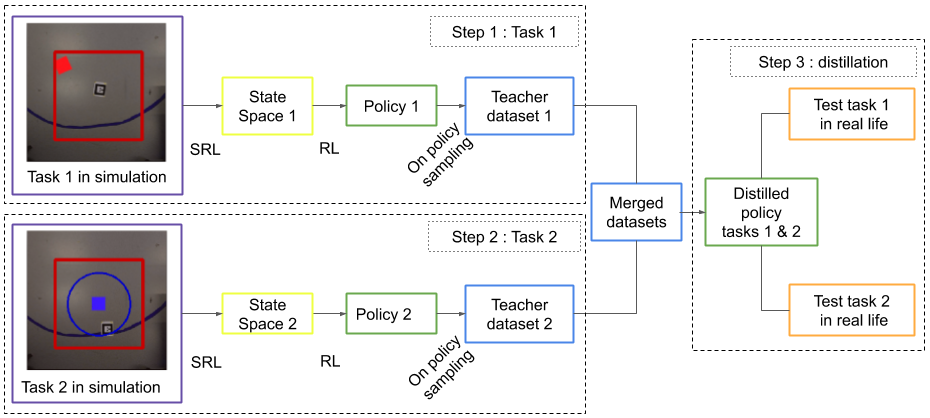}
\caption{Summary of the experimental setup. Step 1 and 2 correspond to learning policies for task 1 and 2, and using those to create distillation datasets. Step 3 is the distillation of the two policies into a single policy which can be deployed in simulation and on the real robot.}
\label{fig:overview}
\end{figure*}

Once the SRL model is trained, we use it's encoder $E_i$ to provide with features for learning a reinforcement learning policy $\pi_i$ with the model $M(\theta)$ ($\theta$ represents the model parameters).
Once $\pi_i$ is learned, we use it to generate sequences of on-policy observations with associated actions, which will eventually be used for distillation (Fig. \ref{fig:overview}, right). We call this the distillation dataset $D_{\pi i}$. We generate $D_{\pi i}$ the following way: we randomly sample a starting position and then let the agent generate a trajectory. At each step we save both the observation and associated action. We stop the sequence when enough reward is gathered (see Section \ref{ref:experiments}).

From each task is only kept the dataset $D_{\pi_i}$. As soon as we change task, $D_{Random,i}$ and $Env_i$ are not available anymore.

In our setting, in order to decrease training time, we generate $D_{Random,i}$ in simulation and learn $\pi_i$ also in simulation. However, at the end of $T$ tasks, $\pi_{D_0,...,D_{T-1}}$ are tested in a real robot. In order to pass the reality gap, the datasets generated are augmented with different luminosity variations.

\subsection{Learning continually}
\label{subsec:continual_learning}

To learn continually we use a distillation algorithm \cite{rusu2015policy}. Once we learned several tasks, we can aggregate the distillation datasets $D_{\pi_i}$ and distill the knowledge into a new model $M_Di(\theta')$ to produce a single flexible 
policy (Fig. \ref{fig:overview}, right). $\theta'$ are the parameters of the distillation model.

The distillation consists in learning in a supervised fashion the action probability associated to an observation at a timestep $t$. Each dataset $D_{\pi_i}$ allows to distill the policy $\pi_i$ into a new network. We name the distilled policy $\pi_{D_i}$. With the aggregation of several distillation datasets, we can distill several policies into the same network. By extension of the previous nomenclature we call a model where policy 1 and policy 2 have been distilled in, $\pi_{D_{1,2}}$. 

At test time, we do not need a task indicator; however, we assume that the observations and state space visually allow to recognize the current task. In the context of continual RL, the task signal is mandatory if the observation does not give any clue about the policy to be run. In our setting, as the policy can be inferred from a different color target tag, we do not need it.

The method presented allows to learn continually several policies without forgetting. On the other hand, $M(\theta)$ also learn on the sequence of task but without any memorization mechanism, its leads to catastrophic forgetting.
The dataset $D_{\pi_i}$ contains 10k samples per task, which allows to learn 
the distillation very quickly (a few minutes are needed to learn $\pi_{D_i}$ while several hours are needed to learn $\pi_i$). 




\subsection{Evaluation}
\label{subsec:evaluation}

\begin{figure*}[ht!]
    \centering
    \includegraphics[scale=0.25]{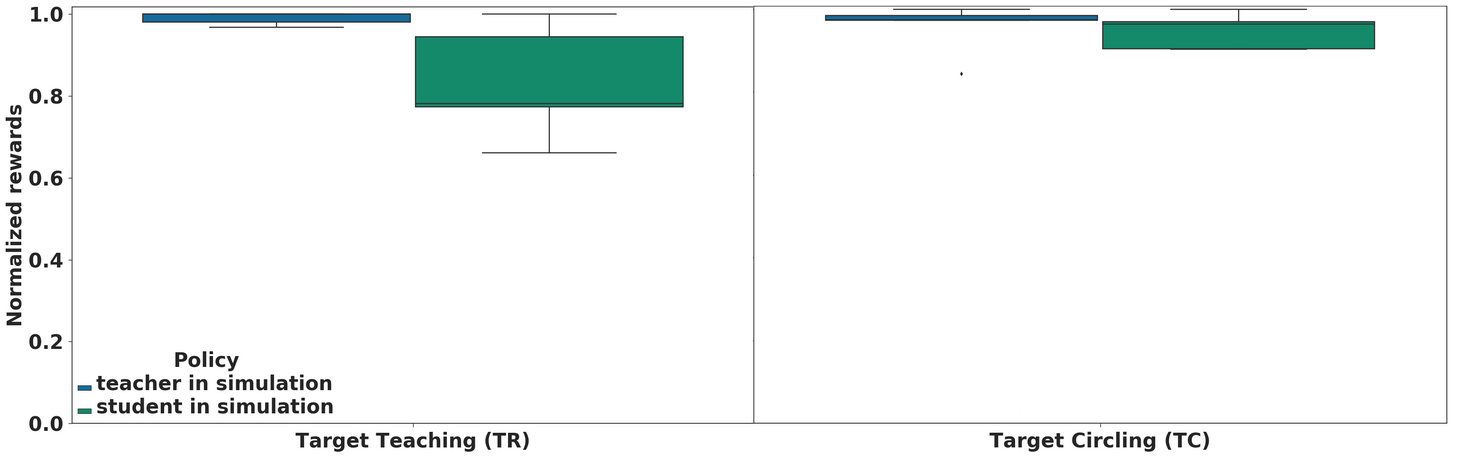} 
    \caption{Comparison between performance (normalized mean reward and standard error) of policy trained on one task only to distilled student policy on the two tasks. The student policy has similar performance on both tasks. \textbf{Left:} Target Reaching (TR). \textbf{Right:} Target Circling (TC) task.}
    \label{fig:final_perf}
\end{figure*}

The main evaluation is the performance of the single and final policy, which can supposedly achieve all previous tasks, as well as being deployed in real life. For that, we report the mean and standard error on 5 runs of the policy on each task in simulation \ref{fig:final_perf}, and provide videos to show the behaviour of the final policy. 

On the other hand we also would like to analyze the learning process.
In order to have an insight on the evolution of the distilled model, we save distillation datasets at different checkpoints in the sequence of tasks. Those checkpoints are saved regularly during the RL training.

%
By distilling and evaluating at several time steps, we are able to evaluate the evolution of learning and forgetting on all environments, both separately and jointly. 
At each checkpoint, we evaluate the actual policy $\pi_i$ on past tasks to assess forgetting and compare it to $\pi_{D_{0,..,t}}$. 

 It is important to note that, even if we consider $Env_i$ as not available anymore at task $i+1$, we did use it for evaluation purposes at any time.
%
 %

\section{Experimental setup}
\label{ref:experiments}

We apply our approach to learn continually two 2D navigation tasks on a real mobile robot.

\subsection{Robotic setup}

The experiments consists of 2D navigation tasks using a 3 wheel omni-directional robot. It is similar to the 2D random target mobile navigation (\cite{raffin2018s}, identical reward setting and possibility of movement). The robot is identified by a black QR code and the scene is recorded from above.

We are able to simulate the experiment, since the robot's input is a fixed RGB image of the scene recorded from above. The robot uses 4 high level discrete actions (move left/right, move up/down in a cartesian plane relative to the robot) rather than motor commands.

The room where the real-life robotic experiments are to be performed is subject to illumination changes. The input image is a top-down view of the floor, which is lighted by surroundings windows and artificial illumination of the room. Hence, the illumination changes depending on the weather and time of the day. We use domain randomization \cite{tobin2017domain} to improve the chances of the policies learned in simulation to better transfer to the real world, by being robust to weather and time conditions. During RL training, at each timestep, the color of the background is randomly changed.

\subsection{Continual learning setup}

Our continual learning scenario is composed of two similar environments, 
where the robot is rewarded according to the associated task. 
In both environments, the robot is free to navigate for up to 250 steps, performing only discrete actions  
within the boundaries identified by a red square.

In environment 1, the robot gets at each timestep $t$ a positive reward +1 for reaching the target identified by a red square marker (task 1), 
a negative reward $R_{t, bump}=-1$ for bumping into the boundaries, and no reward otherwise. 

In environment 2, the robot gets at each timestep $t$ a reward $R_t$ (where $z_t$ is the 2D coordinate position with respect to the center of the circle, see eq. \ref{eq:reward-circular-task}), which is highest when the agent is both keeping a distance to the target equal to a radius $r_{circle}$ (see eq. \ref{eq:reward-cicle}), and has been moving for the previous $k$ steps (see eq. \ref{eq:reward-moving}).
An additional penalty term $R_{t, bump}=-1$ is added to the reward function in case of bump with the boundaries, 
and a coefficient $\lambda=10$ is introduced to balance the behaviour.
$R_t$ is designed for agents to learn the task of circling around a central blue tag (task 2).

\begin{equation}
R_{t, circle} = 1 - \lambda (\|z_t\| -  r_{circle}) ^2  
\label{eq:reward-cicle}
\end{equation}
\begin{equation}
R_{t, movement} = \|z_t -z_{t-k}  \|_{2}^2 
\label{eq:reward-moving}
\end{equation}
\begin{equation}
R_t = \lambda R_{t, circle} *  R_{t, movement} + \lambda ^ 2  R_{t, bump} 
\label{eq:reward-circular-task}
\end{equation}

It is important to note that as the tags associated to each scenario's target are of different color, 
the algorithm can automatically infer which policy it needs to run and thus, does not need task labels at test time.

Moreover, while generating on-policy datasets $D_\pi 1$ (see Section \ref{subsec:oneTask}) for task 1, we allow the robot a limited number of contacts with the target to reach ($N_{contacts}=10$) in order to mainly preserve the frames associated with the correct reaching behaviour. There are no such additional constraints when recording for task 2, the limit is the standard episode size, i.e. 250 time-steps.

The main software related to our experimental setting can be found at the url:  \url{https://github.com/kalifou/robotics-rl-srl/tree/circular_movement_omnibot} \\





\section{Results}

\subsection{Main result}

Our main result is the continual learning of a single policy that solves both tasks in simulation, as presented in Fig. \ref{fig:overview}\footnote{The deployment and evaluation in real life is part of future work}. The two teacher policies are learnt separately (i.e. independently) on each environment. 
Then, distillation is used to combine the two teacher policies into a single policy that can solve the two tasks.

Fig. \ref{fig:final_perf} demonstrates the efficiency of our approach. We can see that the single student distilled policy achieves close to maximum reward in both tasks. 


\subsection{Evaluation of distillation}

\begin{figure}
    \centering
    \includegraphics[scale=0.15]{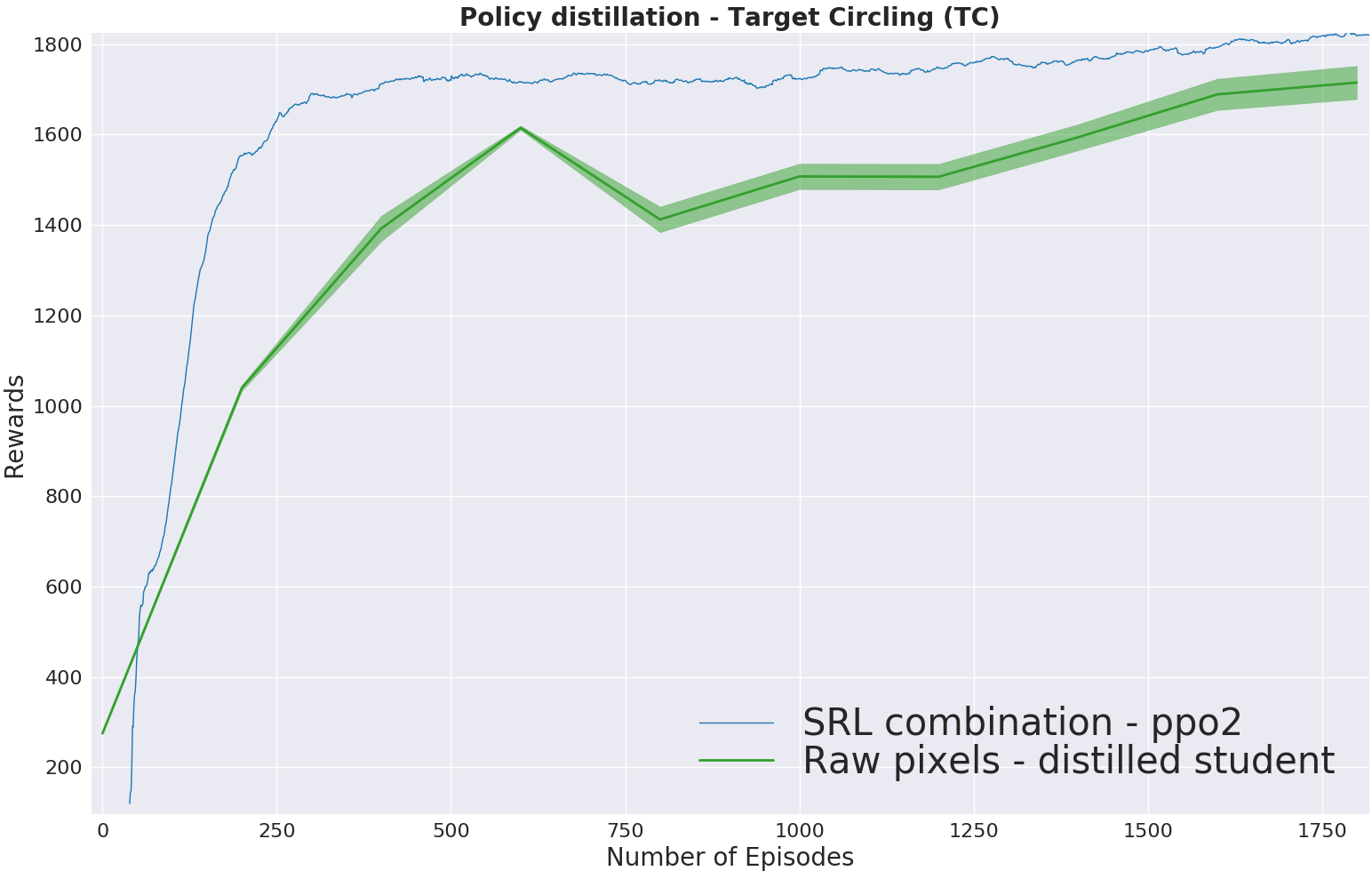}
    \caption{Demonstration of the effectiveness of distillation. Blue: RL training curve of PPO2 on target circling task. Green: Mean and std performance on 8 seeds of distilled student policy.  
    The blue policy is distilled into a student policy at regular time-step (1 episode = 250 timesteps). Both curves are very close, which indicates distillation works as intended.
    }
    \label{fig:distillation_cc}
\end{figure}

We performed a more explicit evaluation of distillation in the task 2 (target circling (TC) around). While we train a policy using RL, we save the policy every 200 episodes (50K timesteps), and distill it into a new student policy which we test. This is illustrated in Fig. \ref{fig:distillation_cc}. Both curves are very close, which indicates distillation works as intended. It is able to transfer a policy using only a limited distillation dataset, with limited loss in the policy performance.

\section{Discussion and future work}
\label{ref:discussion}

Our work is preliminary and offers many possibilities for improvement. Our roadmap include having not only a policy learned in a continual way, but also the SRL model associated. We would need to update the SRL model as new tasks are presented sequentially. One possible approach would be to use Continual SRL methods like S-TRIGGER \cite{caselles2019s} or VASE \cite{achille2018life}.


We also expect to encounter issues when scaling continual learning approaches to more tasks or environments. Indeed, the agent should not accumulate knowledge blindly, but rather make connections between different types of information (i.e. generalize) and/or selectively forget non-useful knowledge. \\

Moreover, we intend to soon provide with supplementary quantitative results and videos of these tasks deployed in the real-life setup. 

We would like to train policies directly on the real robot, as it is the end goal scenario for this research. One promising approach would be to use model-based RL while learning the SRL model
to improve sample efficiency. The final goal would be to learn the policy on a real robot in a reasonable amount of time. 

\section{Conclusion}

In this paper we provide preliminary results towards a proper real life continual learning setup, where a real robot would encounter tasks presented in a sequence and be asked to accumulate knowledge in a scalable manner.
The building blocks for achieving a single policy that solves all presented tasks consists of RL that uses state representation learning models, and distillation into a single policy. This model shows to be a good candidate for transfer to real life and future work should evaluate it in more and more complex tasks.  

\section{Acknowledgement}
This work is supported by the EU H2020 DREAM project (Grant agreement No 640891). 

\newpage

\bibliography{references}
\bibliographystyle{icml2019}

\end{document}